\documentclass[10pt, a4paper]{article}
\usepackage{lrec}
\usepackage{multibib}
\newcites{languageresource}{Language Resources}
\usepackage{graphicx}
\usepackage{tabularx}
\usepackage{soul}

\usepackage{epstopdf}
\usepackage[latin1]{inputenc}
\usepackage{hyperref}
\usepackage{xstring}
\usepackage{times}
\usepackage{latexsym}
\usepackage{url}
\usepackage{color}
\usepackage{subfig}
\usepackage{soul}

\title{Automated email Generation for Targeted Attacks using Natural Language}

\name{Avisha Das, Rakesh Verma}

\address{Department of Computer Science \\
         University of Houston, Houston, Texas \\
         \{adas5, rverma\}@uh.edu\\}

\abstract{
With an increasing number of malicious attacks, the number of people and organizations falling prey to social engineering attacks is proliferating. Despite considerable research in mitigation systems, attackers continually improve their modus operandi by using sophisticated machine learning, natural language processing techniques with an intent to launch successful targeted attacks aimed at deceiving detection mechanisms as well as the victims. 
We propose a system for advanced email masquerading attacks using Natural Language Generation (NLG) techniques. Using legitimate as well as an influx of varying malicious content, the proposed deep learning system generates \textit{fake} emails with malicious content, customized depending on the attacker's intent. The system leverages Recurrent Neural Networks (RNNs) for automated text generation. We also focus on the performance of the generated emails in defeating statistical detectors, and compare and analyze the emails using a proposed baseline. 
\\ \newline \Keywords{natural language generation, email masquerading,
deep learning} }

\begin{document}

\maketitleabstract
 
\section{Introduction}
The continuous adversarial growth and learning has been one of the major challenges in the field of Cybersecurity. With the immense boom in usage and adaptation of the Internet, staggering numbers of individuals and organizations have fallen prey to targeted attacks like phishing and pharming. Such attacks result in digital identity theft causing personal and financial losses to unknowing victims. Over the past decade, researchers have proposed a wide variety of detection methods to counter such attacks (e.g., see \cite{vermaH13,thakurV14,vermaD15,vermaR15,vermaD17}, and references cited therein). However, wrongdoers have exploited cyber resources to launch newer and sophisticated attacks to evade machine and human supervision. Detection systems and algorithms are commonly trained on historical data and attack patterns. Innovative attack vectors can trick these pre-trained detection and classification techniques and cause harm to the victims. 

Email is a common attack vector used by phishers that can be embedded with poisonous links to malicious websites, malign attachments like malware executables, etc~\cite{drake2004anatomy}. 
Anti-Phishing Working Group (APWG) has identified a total of 121,860 unique phishing email reports in March 2017. In 2016, APWG  received over 1,313,771 unique phishing complaints. According to sources in IRS Return Integrity Compliance Services, around 870 organizations had received W-2 based phishing scams in the first quarter of 2017, which has increased significantly from 100 organizations in 2016. And the phishing scenario keeps getting worse as attackers use more intelligent and sophisticated ways of scamming victims.

Fraudulent emails targeted towards the victim may be constructed using a variety of techniques fine-tuned to create the perfect deception. While manually fine-tuning such emails guarantees a higher probability of a successful attack, it requires a considerable amount of time. Phishers are always looking for automated means for launching fast and effective attack vectors. Some of these techniques include bulk mailing or spamming, including action words and links in a phishing email, etc. But these can be easily classified as positive warnings owing to improved statistical detection models. 

Email masquerading is also a popular cyberattack technique where a phisher or scammer after gaining access to an individual's email inbox or outbox can study the nature/content of the emails sent or received by the target. He can then synthesize targeted malicious emails masqueraded as a benign email by incorporating features observed in the target's emails. The chances of such an attack being detected by an automated pre-trained classifier is reduced. The malicious email remain undetected, thereby increasing the chances of a successful attack.  

Current Natural Language Generation (NLG) techniques have allowed researchers to generate natural language text based on a given context. Highly sophisticated and trained NLG systems can involve text generation based on predefined grammar like the Dada Engine~\cite{baki2017scaling} or leverage deep learning neural networks like RNN~\cite{yao2017automated} for generating text. Such an approach essentially facilitates the machine to learn a model that emulates the input to the system. The system can then be made to generate text that closely resembles the input structure and form. 

Such NLG systems can therefore become dangerous tools in the hands of phishers. Advanced deep learning neural networks (DNNs) can be effectively used to generate coherent sequences of text when trained on suitable textual content. Researchers have used such systems for generating textual content across a wide variety of genres - from tweets~\cite{sidhaye2015indicative} to poetry~\cite{ghazvininejad2016generating}. Thus we can assume it is not long before phishers and spammers can use email datasets - legitimate and malicious - in conjunction with DNNs to generate deceptive malicious emails. By masquerading the properties of a legitimate email, such carefully crafted emails can deceive pre-trained email detectors, thus making people and organizations vulnerable to phishing scams.  

In this paper, we address the new class of attacks based on automated fake email generation. We start off by demonstrating the practical usage of DNNs for fake email generation and walk through a process of fine-tuning the system, varying a set of parameters that control the content and intent of the text.   The key contributions of this paper are: 
\begin{enumerate}
\item A study of the feasibility and effectiveness of deep learning techniques in email generation. 
\item Demonstration of an automated system for generation of `fake' targeted emails with a \textit{malicious} intent. 
\item Fine-tuning synthetic email content depending on training data - intent and content parameter tuning.
\item Comparison with a baseline - synthetic emails generated by Dada engine~\cite{baki2017scaling}.
\item Detection of synthetic emails using a statistical detector and investigation of effectiveness in tricking an existing spam email classifier (built using SVM). 
\end{enumerate}

\section{Related Works}
Phishing detection is one of the widely researched areas of cybersecurity. Despite the development of a large number of phishing detection tools, many victims are still falling prey to these attacks. Researchers in ~\cite{drake2004anatomy} explicitly break down the structure of a phishing email, describing in detail the \textit{modus operandi} of a phisher or scammer. In this section, we review previous research in areas of text generation using natural language and the use of deep learning in generation of phishing based attacks and detection.

\textbf{Textual Content Generation.} Natural language generation techniques have been widely popular for synthesizing unique pieces of textual content. NLG techniques proposed by ~\cite{reiter2000building,turner2010generating} rely on templates pre-constructed for specific purposes. The fake email generation system in ~\cite{baki2017scaling} uses a set of manually constructed rules to pre-define the structure of the fake emails. Recent advancements in deep learning networks have paved the pathway for generating creative as well as objective textual content with the right amount of text data for training. RNN-based language models have been widely used to generate a wide range of genres like poetry~\cite{ghazvininejad2016generating,xiedeep}, fake reviews~\cite{yao2017automated}, tweets~\cite{sidhaye2015indicative}, geographical information~\cite{turner2010generating} and many more.

The system used for synthesizing emails in this work is somewhat aligned along the lines of the methodology described in ~\cite{chen2014two,chen2014two2}. However, our proposed system has no manual labor involved and with some level of post processing has been shown to deceive an automated supervised classification system.

\textbf{Phishing email Detection.} In this paper, we focus primarily on generation of fake emails specifically engineered for phishing and scamming victims. Additionally, we also look at some state-of-the-art phishing email detection systems. Researchers in ~\cite{basnet2008detection} extract a large number of text body, URL and HTML features from emails, which are then fed into supervised (SVMs, Neural Networks) as well as unsupervised (K-Means clustering) algorithms for the final verdict on the email nature. The system proposed in~\cite{chandrasekaran2006phishing} extracts 25 stylistic and structural features from emails, which are given to a supervised SVM for analysis of email nature. Newer techniques for phishing email detection based on textual content analysis have been proposed in~\cite{verma2012detecting,vermaH13,vermaA17,Yu2009PhishCatchA}. Masquerade attacks are generated by 
the system proposed in ~\cite{baki2017scaling}, which tunes the generated emails based on legitimate content and style of a famous personality. Moreover, this technique can be exploited by phishers for launching email masquerade attacks, therefore making such a system extremely dangerous.


\section{Experimental Methodology}
The section has been divided into four subsections.  We describe the nature and source of the training and evaluation data in Section~\ref{sec:data} The pre-processing steps are demonstrated in  Section~\ref{sec:dataprep} The system setup and experimental settings have been described in Section~\ref{sec:system}
\subsection{Data description}~\label{sec:data}
To best emulate a benign email, a text generator must learn the text representation in actual legitimate emails. Therefore, it is necessary to incorporate benign emails in training the model. However, as a successful attacker, our main aim is to create the perfect deceptive email - one which despite having malign components like poisoned links or attachments, looks legitimate enough to bypass statistical detectors and human supervision. 

Primarily, for the reasons stated above, we have used multiple email datasets, belonging to both legitimate and malicious classes, for training the system model and also in the quantitative evaluation and comparison steps. For our training model, we use a \textit{larger ratio} of malicious emails compared to legitimate data (approximate ratio of benign to malicious is 1:4). 

\textbf{Legitimate dataset.} 
We use three sets of legitimate emails for modeling our legitimate content. The legitimate emails were primarily extracted from the outbox and inbox of real individuals. Thus the text contains a lot of named entities belonging to PERSON, LOC and ORGANIZATION types. The emails have been extracted from three different sources stated below:
\begin{itemize}
\item 48 emails sent by Sarah Palin (\textbf{Source 1}) and 55 from Hillary Clinton (\textbf{Source 2}) obtained from the archives released in~\cite{palin,clinton} respectively.
\item 500 emails from the Sent items folder of the employees from the Enron email corpus (\textbf{Source 3})~\cite{enron}.
\end{itemize}

\textbf{Malicious dataset.} The malicious dataset was difficult to acquire.  We used two malicious sources of data mentioned below: 
\begin{itemize}
\item 197 Phishing emails collected by the second author - called Verma phish below.
\item 3392 Phishing emails from Jose Nazario's Phishing corpus \footnote{\url{http://monkey.org/~jose/wiki/doku.php} (2004), Deprecated now} (Source 2)
\end{itemize}

\textbf{Evaluation dataset.} We compared our system's output against a small  set of automatically generated emails provided by the authors of~\cite{baki2017scaling}. The provided set consists of 12 emails automatically generated using the Dada Engine and manually generated grammar rules. The set consists of 6 emails masquerading as Hillary Clinton emails and 6 emails masquerading as emails from Sarah Palin. 

Tables~\ref{tab:ldata} and~\ref{tab:pdata} describe some statistical details about the legitimate and malicious datasets used in this system. We define length ($L$) as the number of words in the body of an email. We define Vocabulary ($V$) as the number of unique words in an email. 

\begin{table}[!htb]
\centering
\begin{tabular}{|r|r|r|r|}
\hline
Dataset        & Count & Avg. $L$ & Avg. $V$ \\ \hline
Clinton  & 48    & 32       & 21         \\ \hline
Palin   & 55    & 33       & 26         \\ \hline
Enron   & 500   & 91       & 53         \\ \hline
\textbf{Total}          & 603   & 81       & 48         \\ \hline
\end{tabular}
\caption{Legitimate Data Statistics}
\label{tab:ldata}
\end{table}

\begin{table}[!htb]
\centering
\begin{tabular}{|r|r|r|r|}
\hline
Dataset        & Count & Avg. $L$ & Avg. $V$ \\ \hline
Verma Phish & 197    &  153      & 99          \\ \hline
Nazario Phish   & 3392    &  210      & 129          \\ \hline
\textbf{Total}          & 3589   &  207      & 127          \\ \hline
\end{tabular}
\caption{Phishing Data Statistics}
\label{tab:pdata}
\end{table}
A few observations from the datasets above: the malicious content is relatively more verbose than than the legitimate counterparts. Moreover, the size of the malicious data is comparatively higher compared to the legitimate content.

\subsection{Data Filtering and Preprocessing}~\label{sec:dataprep}
We considered some important steps for preprocessing the important textual content in the data. Below are the common preprocessing steps applied to the data:
\begin{itemize}
\item Removal of special characters like @, \#, \$, \% as well as common punctuations from the email body.
\item emails usually have other URLs or email IDs. These can \textit{pollute} and confuse the learning model as to what are the more important words in the text. Therefore, we replaced the URLs and the email addresses with the $<$LINK$>$ and $<$EID$>$ tags respectively.
\item Replacement of named entities with the $<$NET$>$ tag. We use Python NLTK NER for identification of the named entities.
\end{itemize}

On close inspection of the training data, we found that the phishing emails had incoherent HTML content which can pollute the training model. Therefore, from the original data (in Table~\ref{tab:pdata}), we carefully filter out the emails that were not in English, and the ones that had all the text data was embedded in  HTML. These emails usually had a lot of random character strings - thus the learning model could be \textit{polluted} with such random text. Only the phishing emails in our datasets had such issues.
Table~\ref{tab:pdata_fil} gives the details about the filtered phishing dataset.

\begin{table}[!htb]
\centering
\begin{tabular}{|r|r|r|r|}
\hline
Dataset        & Count & Avg. $L$ & Avg. $V$ \\ \hline
Verma Phish & 127    &  50      & 34          \\ \hline
Nazario Phish   & 2148    &  115      & 71          \\ \hline
\textbf{Total}          & 2275   &  112      & 70          \\ \hline
\end{tabular}
\caption{Phishing Data Statistics after filtering step}
\label{tab:pdata_fil}
\end{table}

\subsection{Experimental Setup}~\label{sec:system}
We use a deep learning framework for the Natural Language Generation model. The system used for learning the email model is developed using Tensorflow 1.3.0 and Python 3.5. This section provides a background on a Recurrent Neural Network for text generation. 

Deep Neural Networks are complex models  for computation with deeply connected networks of neurons to solve complicated machine learning tasks. Recurrent Neural Networks (RNNs) are a type of deep learning networks better suited for sequential data. RNNs can be used to learn  character and word sequences from natural language text (used for training). The RNN system used in this paper is capable of generating text by varying levels of granularity, i.e. at the character level or word level. For our training and evaluation, we make use of Word-based RNNs since previous text generation systems~\cite{xiedeep}, \cite{henderson2014word} have generated coherent and readable content using word-level models. A comparison between Character-based and Word-based LSTMs in ~\cite{xiedeep} proved that for a sample of generated text sequence, word level models have lower perplexity than character level deep learners. This is because the character-based text generators suffer from spelling errors and incoherent text fragments.  
\subsubsection{RNN architecture}
Traditional language models like N-grams are limited by the history or the sequence of the textual content that these models are able to look back upon while training. However, RNNs are able to retain the long term information provided by some text sequence, making it work as a ``memory''-based model. However while building a model, RNNs are not the best performers when it comes to preserving long term dependencies. For this reason we use Long Short Term Memory architectures (LSTM) networks which are able to learn a better language/text representation for longer sequences of text. 

We experiment with a few combinations for the hyper-parameters- number of RNN nodes, number of layers, epochs and time steps were chosen empirically. The input text content needs to be fed into our RNN in the form of word embeddings. The system was built using 2 hidden LSTM layers and each LSTM cell has 512 nodes. The input data is split into mini batches of 10 and trained for 100 epochs with a learning rate of $2 \times 10^{-3}$. The sequence length was selected as 20. We use $cross-entropy$ or $softmax$ optimization technique~\cite{Goodfellow-et-al-2016} to compute the training loss, $Adam$ optimization technique~\cite{Kingma2014AdamAM} is used for updating weights. The system was trained on an Amazon Web Services EC2 Deep Learning instance using an Nvidia Tesla K80 GPU. The training takes about 4 hours. 

\subsubsection{Text Generation and Sampling}
The trained model is used to generate the email body based on the nature of the input. We varied the sampling technique of generating the new characters for the text generation. 

\textbf{Generation phase.} Feeding a word ($\widehat{w_{0}}$) into the trained LSTM network model, will output the word most likely to occur after $\widehat{w_{0}}$ as $\widehat{w_{1}}$ depending on $P(\widehat{w_{1}}~|~\widehat{w_{0}})$. If we want to generate a text body of $n$ words, we feed $\widehat{w_{1}}$ to the RNN model and get the next word by evaluating $P(\widehat{w_{2}} ~|~\widehat{w_{0}},\widehat{w_{1}})$. This is done repeatedly to generate a text sequence with n words: $\widehat{w_{0}}$, $\widehat{w_{1}}$, $\widehat{w_{2}}$, ..., $\widehat{w_{n}}$.   

\textbf{Sampling parameters.} We vary our sampling parameters to generate the email body samples. For our implementation, we choose \textit{temperature} as the best parameter. Given a sequence of words for training, $w_{0}$, $w_{1}$, $w_{2}$, ..., $w_{n}$, the goal of the trained LSTM network is to predict the best set of words that follow the training sequence as the output ($\widehat{w_{0}}$, $\widehat{w_{1}}$, $\widehat{w_{2}}$, ..., $\widehat{w_{n}}$). 

Based on the input set of words, the model builds a probability distribution 
$P(w_{t+1} ~|~ w_{t'\leq t}) = softmax(\widehat{w_{t}})$, 
here $softmax$ normalization with \textit{temperature} control (Temp) is defined as:

$P(softmax(\widehat{w_{t}^{j}})) = \frac{K(\widehat{w_t^j}, Temp)}{\sum_{j=1}^{n} K(\widehat{w_t^j}, Temp)} $, where $K(\widehat{w_t^j}, Temp) = e^{\frac{\widehat{w_{t}^{j}}}{Temp}}$

The novelty or eccentricity of the RNN text generative model can be evaluated by varying the Temperature parameter  between $0 < Temp. \leq 1.0$ to generate samples of text (the maximum value is 1.0). We vary the nature of the model's predictions using two main mechanisms - deterministic and stochastic. Lower values of $Temp.$ generates relatively deterministic samples while higher values can make the process more stochastic. Both the mechanisms suffer from issues, deterministic samples can suffer from repetitive text while the samples generated using the stochastic mechanism are prone to spelling mistakes, grammatical errors, nonsensical words.  
We generate our samples by varying the temperature values to 0.2, 0.5, 0.7 and 1.0. For our evaluation and detection experiments, we randomly select 25 system generated samples, 2 samples generated at a temperature of 0.2, 10 samples at temperature 0.5, 5 samples at a temperature of 0.7 and 8 samples at temperature 1.0.   

\subsubsection{Customization of Malicious Intent} 
One important aspect of malicious emails is their harmful intent. 
The perfect attack vector will have malicious elements like a poisonous link or malware attachment wrapped in legitimate context, something which is sly enough to fool both a state-of-the-art email classifier as well as the victim. One novelty of this system training is the procedure of \textbf{injecting} malicious intent during training and \textbf{generating} malicious content in the synthetic emails.

We followed a \textit{percentage based influx} of malicious content into the training model along with the legitimate emails. The training models were built by varying the percentage (5\%, 10\%, 30\% and 50\%) of phishing emails selected from the entire phishing dataset along with the entire legitimate emails dataset. We trained separate RNN models on all these configurations. For studying the varying content in emails, we generate samples using temperature values at 0.2, 0.5, 0.7 and 1.0.

\subsection{Detection using Existing Algorithms}
We perform a simple quantitative evaluation by using three text-based classification algorithms on our generated emails. Using the Python SciKit-Learn library, we test three popular text-based filtering algorithms - Support Vector Machines~\cite{maldonado2013svm}, Naive Bayes~\cite{witten2016data} and Logistic Regression~\cite{franklin2005elements}. 

The training set was modeled as a \textit{document-term matrix} and the \textit{word count vector} is used as the feature for building the models. For our evaluation, we train models using Support Vector Machines (SVM), Naive Bayes (NB) and Logistic Regression (LR) models on a training data of 300 legitimate emails from WikiLeaks archives\footnote{\url{https://wikileaks.org/}} and 150 phishing emails from Cornell PhishBowl~\cite{cornellphish}. We test the data on 100 legitimate emails from WikiLeaks archives that were not included in the training set and 25 `fake' emails that were generated by our natural language generation model. 

\section{Analysis and Results}~\label{sec:results}
We discuss the results of the generative RNN model in this section. We give examples of the email text generated with various training models and varying temperatures. We also provide the accuracy of the trained classifiers on a subset of these generated email bodies (after slight post processing). We try to provide a qualitative as well as a quantitative review of the generated emails.
\subsection{Examples of Machine generated emails}~\label{sec:examples}
\textbf{(A) \textit{Training only on Legitimates and varying sampling temperatures}} \\
We show examples of emails generated using models trained on legitimate emails and sampled using a temperature of 1.0.\\

\textbf{Example I at Temperature = 1.0:}\\
\fbox{\begin{minipage}{23em}
Dear $<$NME$>$ The article in the $<$NME$>$  offers promotion should be somewhat changed for the next two weeks. $<$NME$>$ See your presentation today. $<$NME$>$ 
\end{minipage}}
\\
\eject
\textbf{Example II Example I at Temperature = 0.7:}\\
\fbox{\begin{minipage}{23em}
Sir I will really see if they were more comments tomorrow and review
and act upon this evening $<$NET$>$. The engineer I can add
there some $<$LINK$>$ there are the issues $<$NET$>$. Could you give me a basis for
the call \textcolor{red}{him he said}
\end{minipage}}\\
The example above shows that while small substrings make some sense. The sequence of text fragments generated make very little sense when read as a whole. When comparing these with the phishing email structure described in ~\cite{drake2004anatomy}, the generated emails have very little malicious content. The red text marks the incongruous text pieces that do not make sense. \\
\textbf{(B) \textit{Training on Legitimates + 5\% Malicious content:} }\\
In the first step of intent injection, we generate emails by providing the model with all the legitimate emails and 5\% of the cleaned phishing emails data (Table~\ref{tab:pdata_fil}). Thus for this model, we create the input data with 603 legitimate emails and 114 randomly selected phishing emails. We show as examples two samples generated using temperature values equal to 0.5 and 0.7.  \\

\textbf{Example I at Temperature = 0.5:}\\
\fbox{\begin{minipage}{23em}
Sir Here are the above info on a waste of anyone, but an additional figure and it goes to $<$NET$>$. Do I $<$NET$>$ got the opportunity for a possible position between our Saturday $<$NME$>$ or $<$NET$>$ going to look over you in a presentation you will even need $<$NET$>$ to drop off the phone.
\end{minipage}}\\

\textbf{Example II at Temperature = 0.7:}\\
\fbox{\begin{minipage}{23em}
Hi owners $<$NET$>$ your Private $<$NET$>$ email from $<$NET$>$ at 
$<$NET$>$ email $<$NET$>$ \textcolor{blue}{Information I'll know our pending your fake check
to eol} thanks $<$NET$>$ and would be  In maintenance in a long
online demand
\end{minipage}}

The model thus consists of benign and malicious emails in an approximate ratio of 5:1. Some intent and urgency can be seen in the email context. But the incongruent words still remain.\\

\textbf{(C) \textit{Training on Legitimates + 30\% Malicious content:} }\\
We further improve upon the model proposed in (B). In this training step, we feed our text generator all the legitimate emails (603 benign) coupled with 30\% of the malicious emails data (683 malicious). This is an almost balanced dataset of benign and phishing emails. The following examples demonstrate the variation in text content in the generated emails.  \\

\textbf{Example I at Temperature = 0.5:}\\
\fbox{\begin{minipage}{23em}
Sir we account access will do so may not the emails about the $<$NET$>$ This $<$NET$>$ is included at 3 days while when to $<$NET$>$ because \textcolor{blue}{link below to update your account until the deadline} we will received this information that we will know that your $<$NET$>$ account information needs
\end{minipage}}\\
\eject
\textbf{Example II at Temperature = 1.0:}\\
\fbox{\begin{minipage}{23em}
Dear registered \textcolor{red}{secur= online}, number: hearing from This trade guarded please account go to pay it. To \textcolor{blue}{modify your Account then fill in necessary from your notification preferences,} please PayPal account provided with the integrity of information on the Alerts tab. 
\end{minipage}}

A good amount of text seems to align with the features of malicious emails described in~\cite{drake2004anatomy} have malicious intent in it. We choose two examples to demonstrate the nature of text in the generated emails. We include examples from further evaluation of steps.\\

\textbf{(D) \textit{Training on Legitimates + 50\% Malicious content:} }\\

In this training step, we consider a total of 50\% of the malicious data (1140 phishing emails) and 603 legitimate emails. This is done to observe whether training on an unbalanced data, with twice the ratio of malign instances than legitimate ones, can successfully incorporate obvious malicious flags like poisonous links, attachments, etc. We show two examples of emails generated using deep learners at varying sampling temperatures. \\

\textbf{Example I at Temperature = 0.7:}\\
\fbox{\begin{minipage}{23em}
If you are still online. genuine information in the message, notice your account has been frozen to your account \textcolor{blue}{in order to restore your account as click on CONTINUE Payment Contact $<$LINK$>$ UK}.
\end{minipage}} \\

\textbf{Example IT at Temperature = 0.5:}\\
\fbox{\begin{minipage}{23em}
Hi will have temporarily information your account will be restricted
during that the Internet accounts and upgrading password An data
Thank you for your our security of your Account \textcolor{blue}{Please click on it using
the $<$NET$>$ server} This is an new offer miles with us as a qualified and
move in
\end{minipage}}\\

The generated text reflects malicious features like URL links and tone of urgency. We can assume that the model picks up important cues of malign behavior. The model then learns to incorporate such cues into the sampled data  during training phase.


\subsection{Evaluation using Detection Algorithm}
We train text classification models using Support Vector Machines (SVM), Naive Bayes (NB) and Logistic Regression (LR) models on a training data of 300 legitimate emails from WikiLeaks archives\footnote{\url{https://wikileaks.org/}} and 150 phishing emails from Cornell PhishBowl~\cite{cornellphish}. We test the data on 100 legitimate emails from WikiLeaks archives that were not included in the training set and 25 `fake' emails that were generated by our natural language generation model trained on a mix of legitimate and 50\% malicious emails. We randomly select the emails (the distribution is: 2 samples generated at a temperature of 0.2, 10 samples at temperature 0.5, 5 samples at a temperature of 0.7 and 8 samples at temperature 1.0) for our evaluation. 

We use the Scikit-Learn Python library to generate the \textit{document-term matrix} and the \textit{word count vector} from a given sample of email text body used as a feature for training the classification models. 
The Table~\ref{tab:res} reports the accuracy, precision, recall, and F1-scores on the test dataset using SVM, Naive Bayes and Logistic Regression classifiers.

\begin{table}[!htb]
\centering
\begin{tabular}{|l|l|l|l|l|}
\hline
Classifier & Accuracy & Precision & Recall & F1-score \\\hline 
SVM & 71&72&85&78 \\\hline 
NB &78&91&75&82 \\\hline 
LR &91&93&95&94 \\\hline 
\end{tabular}
\caption{Classification metrics of generated emails}
\label{tab:res}
\end{table}

Despite the incoherent nature of the generated emails, the text-based classifiers do not achieve a 100\% accuracy as well as F1-scores. 

\subsection{Comparison of emails with another NLG model}
The authors in ~\cite{baki2017scaling} discuss using a Recursive Transition Network for generating fake emails similar in nature to legitimate emails. The paper discusses a user study testing the efficacy of these fake emails and their effectiveness in being used for deceiving people. The authors use only legitimate emails to train their model and generate emails similar to their training data - termed as `fake' emails. In this section, we compare a couple of examples selected randomly from the emails generated by the Dada Engine used in \cite{baki2017scaling} and the outputs of our Deep Learning system  generated emails. \\

\textbf{Generated by the RNN (Example I):}\\
\fbox{\begin{minipage}{23em}
Hi will have temporarily information your account will be restricted
during that the Internet accounts and upgrading password An data
Thank you for your our security of your Account \textcolor{blue}{Please click on it using
the $<NET>$ server} This is an new offer miles with us as a qualified and
move in\\

\end{minipage}}
\newline\\
\textbf{Generated by the RNN (Example II):}\\
\fbox{\begin{minipage}{23em}
Sir Kindly limit, it \textcolor{red}{[IMAGE]} Please contact us contained on this suspension will not be \textcolor{red}{=} interrupted by \textcolor{red}{10} product, or this \textcolor{red}{temporary cost some of the} 
\end{minipage}}

\textbf{Generated by the Dada Engine:}\\
\fbox{\begin{minipage}{23em}
Great job on the op-ed! Are you going to submit? Also, Who will be
attending?
\end{minipage}}

The examples provide evidence that emails generated by the RNN are more on the lines of phishing emails than the emails generated by the Dada Engine. Of course, the goal of the email generated by the Dada engine is masquerade, not phishing. Because of the rule-based method employed that uses complete sentences, the emails generated by the Dada engine have fewer problems of coherence and grammaticality. 
\section{Error Analysis}~\label{sec:error}
We review \textit{two types of errors} observed in the evaluation of our RNN text generation models developed in this study. \textit{First}, the text generated by multiple RNN models suffer from repetitive tags and words. The example of the email body below demonstrates an incoherent and absurd piece of text generated by the RNN trained on legitimate emails and 50\% of phishing emails with a temperature of 0.5.\\

\fbox{\begin{minipage}{23em}
Hi 48 PDX Cantrell $<$LINK$>$ \textcolor{red}{$<$NET$>$ $<$NET$>$ ECT ECT $<$NET$>$ $<$NET$>$ ECT ECT $<$NET$>$ $<$NET$>$ ECT ECT $<$NET$>$ $<$NET$>$ ECT ECT $<$NET$>$ F $<$NET$>$ ECT ECT $<$NET$>$ G Slaughter 06 07 03 57 DEVELOPMENT 06 09 2000 07 01 $<$NET$>$ $<$NET$>$ ECT} ENRON 09 06 03 10 23 PM To \textcolor{red}{$<$NET$>$ $<$NET$>$ ECT ECT cc $<$NET$>$ $<$NET$>$ ECT ECT} Subject Wow Do not underestimate the employment group contains Socal study about recession impact $<$NET$>$ will note else to you for a revised Good credit period I just want to bring the afternoon $<$NET$>$ I spoke to $<$NET$>$ Let me know if
\end{minipage}}
\newline \\

This kind of repetitive text generation was observed a number of times. However, we have not yet investigated the reasons for these repetitions. This could be an inherent problem of the LSTM model, or it could be because of the relatively small training dataset we have used. A third issue could be the temperature setting. More experiments are needed to determine the actual causes.  

The \textit{second aspect} of error analysis is to look at the misclassification by the statistical detection  algorithms. Here we look at a small sample of emails that were marked as legitimate despite being fake in nature. We try to investigate the factors in the example sample that can explain the misclassification errors by the algorithms.\\

\textbf{Example (A):} \\
\fbox{\begin{minipage}{23em}
Hi GHT location $<$EID$>$ Inc Dear $<$NET$>$ Password Location $<$NET$>$ of $<$NET$>$ program We have been riding to meet In a of your personal program or other browser buyer buyer The email does not commit to a secure F or security before You may read a inconvenience during Thank you $<$NET$>$
\end{minipage}}
\newline \\

\textbf{Example (B):} \\
\fbox{\begin{minipage}{23em}
Sir we account access will do so may not the emails about the $<$NET$>$ This $<$NET$>$ is included at 3 days while when to $<$NET$>$ because the link below to update your account until the deadline we will received this information that we will know that your $<$NET$>$ account information needs
\end{minipage}}
\newline \\

\textbf{Example (C):} \\
\fbox{\begin{minipage}{23em}
Sir This is an verificati= $<$LINK$>$ messaging center, have to inform you that we are conducting more software,  Regarding Your Password : $<$LINK$>$ \textcolor{red}{\& June 20, 2009 Webmail} Please Click Here to Confirm
\end{minipage}}
\newline \\

Examples (A), (B) and (C) are emails generated from a model trained on legitimate and 50\% of phishing data (Type (D) in Section~\ref{sec:examples}) using a temperature of 0.7. There can be quite a few reasons for the misclassification - almost all the above emails despite being `fake' in nature have considerable overlap with words common to the legitimate text. Moreover, Example (A) has lesser magnitude of indication of malicious intent. And the amount of malicious intent in Example (B), although notable to the human eye, is enough to fool a simple text-based email classification algorithm. Example (C) has multiple link tags implying possible malicious intent or presence of poisonous links. However, the position of these links play an important role in deceiving the classifier. A majority of phishing emails have links at the end of the text body or after some action words like \textit{click}, \textit{look}, \textit{here}, \textit{confirm} etc. In this case, the links have been placed at arbitrary locations inside the text sequence - thereby making it harder to detect.   These misclassification or errors on part of the classifier can be eliminated by human intervention or by designing a more sensitive and sophisticated detection algorithm.

\section{Conclusions and Future Work}
While the RNN model generated text which had `some' malicious intent in them - the examples shown above are just a few steps from being coherent and congruous. We designed an RNN based text generation system for generating targeted attack emails which is a challenging task in itself and a novel approach to the best of our knowledge. The examples generated however suffer from random strings and grammatical errors. We identify a few areas of improvement for the proposed system - reduction of repetitive content as well as inclusion of more legitimate and phishing examples for analysis and model training. We would also like to experiment with addition of topics and tags like `bank account', `paypal', `password renewal', etc. which may help generate more specific emails. It would be interesting to see how a generative RNN handles topic based email generation problem.


\section{Bibliographical References}
\label{main:ref}

\bibliographystyle{lrec}
\bibliography{xample}


\end{document}